\documentclass{article}

\usepackage[nonatbib, preprint]{nips_2018}

\usepackage[utf8]{inputenc} 
\usepackage[T1]{fontenc}    
\usepackage{hyperref}       
\usepackage{url}            
\usepackage{booktabs}       
\usepackage{amsfonts}       
\usepackage{nicefrac}       
\usepackage{microtype}      
\usepackage{amsmath}
\usepackage{graphicx}
\graphicspath{ {figures/} }
\usepackage{subcaption}
\usepackage{cite}
\usepackage{threeparttable}

\title{SSIMLayer: Towards Robust Deep Representation Learning via Nonlinear Structural Similarity}

\author{
	Ahmed Abobakr, Mohammed Hossny and Saeid Nahavandi\\
	Institute for Intelligent Systems Research and Innovation (IISRI)\\
	Deakin University\\
	Australia \\
	\texttt{\{aabobakr, mo.hossny, saeid.nahavandi\}@deakin.edu.au} \\
}

\begin{document}
	
	\maketitle
	
	\begin{abstract}
		Deeper convolutional neural networks provide more capacity to approximate complex mapping functions. However, increasing network depth imposes difficulties on training and increases model complexity. This paper presents a new nonlinear computational layer of considerably high capacity to the deep convolutional neural network architectures. This layer performs a set of comprehensive convolution operations that mimics the overall function of the human visual system (HVS) via focusing on learning structural information in its input. The core of its computations is evaluating the components of the structural similarity metric (SSIM) in a setting that allows the kernels to learn to match structural information. The proposed SSIMLayer is inherently nonlinear and hence, it does not require subsequent nonlinear transformations. Experiments conducted on CIFAR-10 benchmark demonstrates that the SSIMLayer provides better convergence than the traditional convolutional layer, bypasses the need for nonlinear transformations and shows more robustness against noise perturbations and adversarial attacks.
	\end{abstract}
	
	\section{Introduction}
	
	Deep representation learning architectures have achieved superior perceptual capabilities in several domains. In particular, the deep convolutional neural network (CNN) has dominated complex visual perception tasks such as object recognition~\cite{he2016}, object detection and localisation~\cite{he2017mask}, and semantic segmentation~\cite{li2017fcis, he2017mask}. The CNN provides superior learning capacity to approximate complex mapping functions via a stack of computational layers that is based on the linear convolution operator~\cite{lecun2015}. This stack is trained end-to-end using general purpose gradient optimisation algorithms to extract features with an increasing level of abstraction~\cite{lecun2015}. The importance of building deeper models has been demonstrated in several studies to provide more powerful learning capabilities~\cite{he2016}. However, deeper architectures are difficult to be optimised due to inherited problems such as vanishing and exploding gradients~\cite{he2016}. While the residual learning paradigm~\cite{he2016} has been successful in mitigating the effect of unstable gradients, this remains an open research problem for deep architectures. Further, the runtime complexity that may result from deeper models limits deployment on embedded devices.
	
\begin{figure}[h]
		{\includegraphics[width=\linewidth]{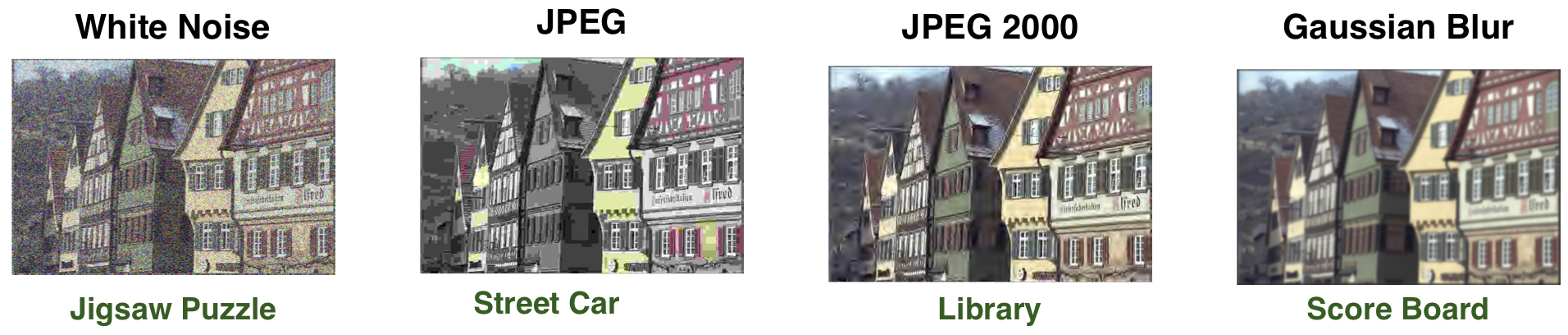}}
		\caption[The effect of different noise models on the predictions of a ConvNet model]
		{The effect of different noise models and lossy compression techniques on the predictions of a ConvNet model. The model produced different predictions for different kinds of distortion.} 
		\label{fig:noise}
	\end{figure}

	Sensitivity to noise and input distortions is another issue that challenges deep machine learning models, especially the CNN models. These models have a major limitation in understanding and eliminating the effect of noise. It has been demonstrated that imperceptible perturbations can dramatically change the outcome of a CNN model~\cite{goodfellow2014explaining, moosavi2016deepfool,krizhevsky2012}. We have performed several experiments to study the effect of noise on these architectures. Images are augmented with different noise models and fed to the popular AlexNet model~\cite{krizhevsky2012}. As shown in Fig.~\ref{fig:noise}, the response of AlexNet model changes according to the type and strength of the added noise.
	
\begin{figure}[h]
		\centering
		\includegraphics[width=\linewidth]{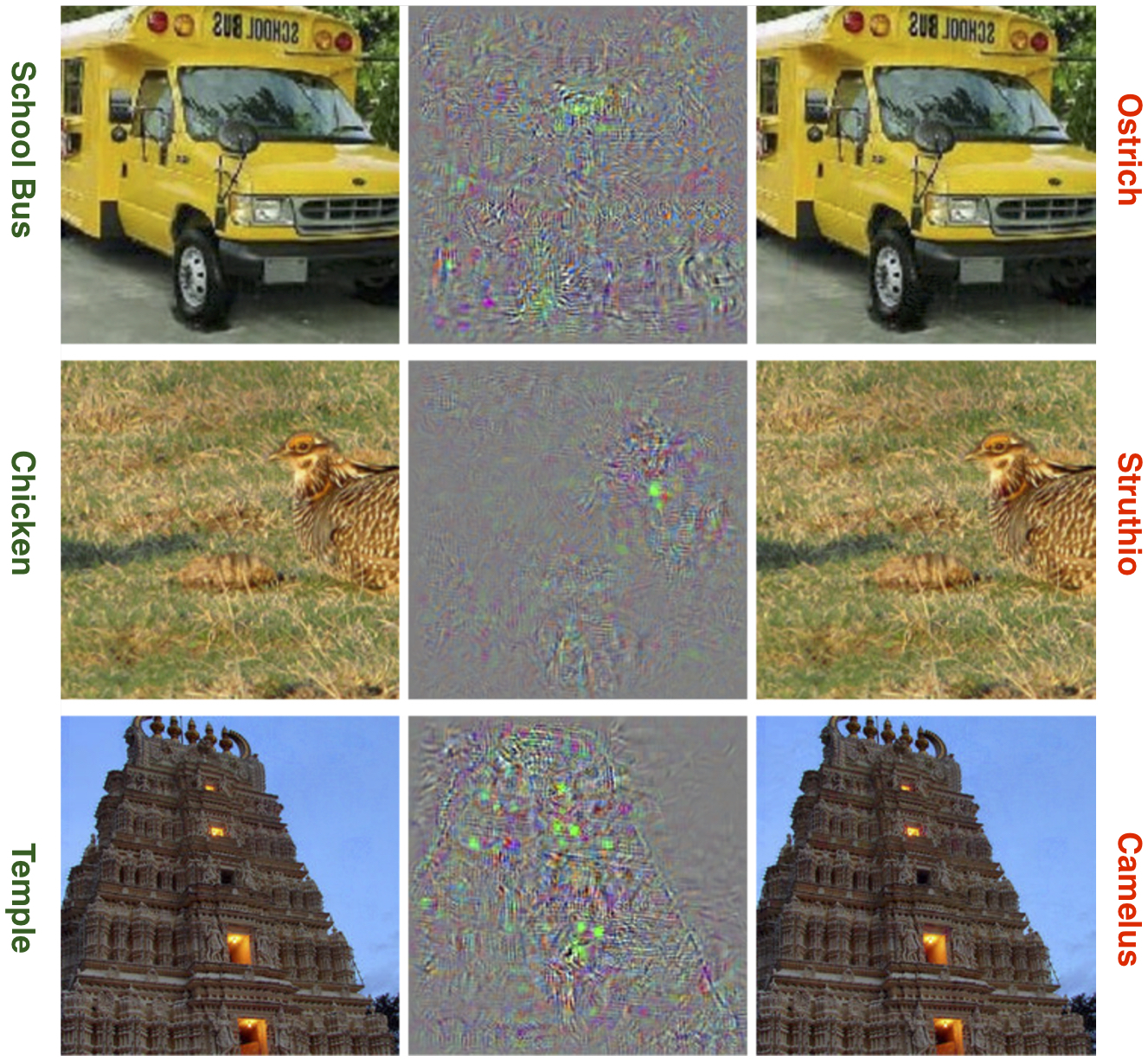}
		\caption[An unobservable distortion totally changes the predictions of AlexNet]{An unobservable distortion totally changes the predictions of AlexNet~\cite{krizhevsky2012}. Left column is the correctly predicted images, right is the distorted incorrectly classified samples, center column is the difference between the original and distorted images. Figure from~\cite{szegedy2013intrig}.}
		\label{fig:s13noise}
	\end{figure}

\begin{figure}[h]
		{\includegraphics[width=\linewidth]{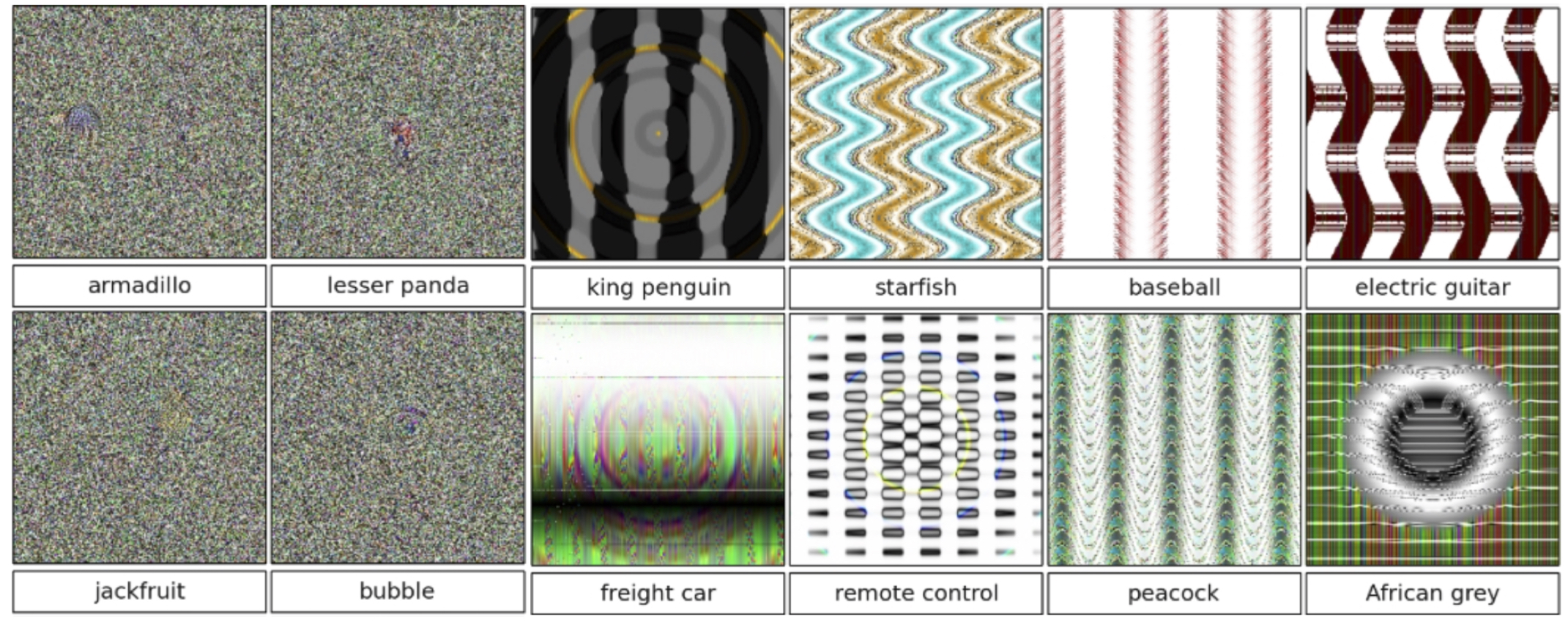}}
		\caption[Deep neural networks are easily fooled]{Deep neural networks are easily fooled. AlexNet~\cite{krizhevsky2012}, is highly confident with unrecognisable images. Figure from~\cite{nguyen2015}.} 
		\label{fig:fooled}
	\end{figure}

	Further investigations were performed by other researchers~\cite{szegedy2013intrig,nguyen2015}. In\cite{szegedy2013intrig}, adding an optimised imperceptible distortion to an image led to a totally different prediction by the deep network, see Fig.~\ref{fig:s13noise}. Recently, Nguyen et al.~\cite{nguyen2015}, revealed that deep neural networks are easily fooled, using different approach. In their work, an evolutionary algorithm is used to find a set of images that are predicted with a high confidence level by AlexNet. They have found that the deep network is highly confident with totally unrecognisable images, as shown in Fig.~\ref{fig:fooled}. 
	
	\bigbreak
	To that end, this paper proposes the SSIMLayer. A new nonlinear computational layer of high learning capacity to the deep convolutional neural network architectures. The SSIMLayer is perceptually inspired and designed to incorporate the HVS functionalities into the deep CNN models. The HVS is a complex nonlinear system that is highly adapted for extracting structural information of the visual world. CNN architectures are trying to achieve the human level of visual perception via learning a hierarchy of features from raw input. However, the simple linear convolution operator is not sufficient for a neuron to extract structural information. Therefore, the proposed formulation adapts the CNN neurons to extract structural information from input images and disregard illumination and contrast effects. 
	\smallbreak
	The motivation for the SSIMLayer is three fold. First, the proposed layer iherently incorporates the functionalities of the HVS into CNN architectures. Second, the SSIM measure is nonlinear and differentiable, hence, its parameters can be optimised using backpropagation. Third, focusing on extracting structural information can help reduce model complexity via building shallower and more powerful models overcoming gradients instability problems~\cite{he2016, ioffe2015batch}.   
	\smallbreak
	In the literature, the SSIM metric has received extensive research interest from the deep learning community as it outperforms traditional objective image assessment metrics in quantifying the quality of perceived images, and satisfies the differentiability requirement for the backpropagation stage. It has mainly been used as a loss function in optimising unsupervised deep generative models. These models try to learn a compact representation from unlabeled training images through minimising the reconstruction error of the input. Generative adversarial networks (GAN) and auto-encoders are the most widely used generative models. Unlike the mean squared reconstruction error metric, the SSIM is well matched with the perceived visual quality and  makes use of the strong local dependencies of pixels~\cite{wang2004}. Zhao et al.~\cite{zhao2017loss} provides a review of loss functions used with deep neural networks and demonstrates the superiority of the perceptually inspired SSIM loss and its variants.

	\section{Structural Similarity Index SSIMLayer}
	\label{sc:ssim}
	The SSIM~\cite{wang2004} is a full-reference objective image quality assessment metric. It has been formulated under the assumption that the HVS is highly adapted for extracting structural information from a visual input. The SSIM index quantifies the degradation of structural information between a distorted and its corresponding reference image. Moreover, the SSIM metric is nonlinear and its operations are differentiable satisfying the requirements for backpropagation and gradient based optimisation techniques. Therefore, in this work, the formulation of the SSIM metric is adjusted and incorporated as a computational layer in deep learning architectures.
	
	\smallbreak
	
	Neurons of the SSIMLayer perform more expressive convolution operations that aim at comparing structural similarity independent from luminance and contrast. During training, layer filters are updated using backprobagated gradients to jointly maximise the structural similarity with spatial local patches in the input and minimise the training loss function. This yields an average structural memory image that represents dominant structures in the training dataset. The firing rate of neurons is controlled with the degree of similarity between trained filters and local input patches that have been normalised for luminance and contrast. Therefore, the final outcome is an SSIM activation map where each component is a combination of three comparisons; luminance ($l$), contrast ($c$) and structure ($s$), representing the degree of structural similarity between two aligned local input patch and filter. 
	
	\subsection{Mathematical Formulation}
	
	Given two aligned local input patch $x$ and a trainable weight filter $y$, the SSIM activation is computed as:
	\begin{align}
	\text{SSIM}(x,y) &= [l(x,y)]^{\alpha}~.~[c(x,y)]^\beta~.~[s(x,y)]^\gamma, \\
	l(x,y) &= \frac{2\mu_{x}\mu_{y} + C1}{\mu_{x}^{2} + \mu_{y}^{2} + C1 },	 \\
	\mu_{x} &= \frac{1}{N} \sum_{i=1}^{N} x_{i}, \\
	c(x,y) &= \frac{2\sigma_{x}\sigma_{y} + C2}{\sigma_{x}^{2} + \sigma_{y}^{2} + C2 }, \\
	\sigma_{x} &= \bigg( \frac{1}{N - 1}  \sum_{i=1}^{N} (x_{i} - \mu_{x})^2 \bigg)^\frac{1}{2}, \\
	s(x,y) &= \frac{2\sigma_{xy} + C3}{\sigma_{x} \sigma_{y} + C3 }, \\
	\sigma_{xy} &= \frac{1}{N - 1}  \sum_{i=1}^{N} (x_{i} - \mu_{x})(y_{i} - \mu_{y})
	\label{eq:ssimform}
	\end{align}
	where $\alpha > 0,\beta > 0, \gamma > 0$ are parameters denoting the importance of the relative component and $C1, C2$ and $C3$ are added constants to ensure numerical stability. Setting $\alpha = \beta = \gamma = 1 ~\text{and}~C_{3} = C_{2} / 2$ simplifies the computations to: 
	\begin{equation}
	\text{SSIM}(x,y) = \frac{(2\mu_{x}\mu_{y} + C_{1})(2\sigma_{xy} + C_{2})}{(\mu_{x}^{2} + \mu_{y}^{2} + C_{1})(\sigma_{x}^{2} + \sigma_{y}^{2} + C2)}.
	\label{eq:ssim}
	\end{equation}
	
	\subsection{Gradient Based Learning for SSIM Parameters}
	\label{sec:ssim_grads}
	
	The SSIM is differentiable, hence it satisfies the main requirement for the backpropagation stage. Parameters of the SSIMLayer are optimised using gradient based optimisation techniques. In this approach, weights of neurons are iteratively adjusted to jointly maximise the structural similarity with the local patch in the input and minimise a global performance measure. The gradient descent procedure converges to a local minima, which is most probably close to the global minima~\cite{lecun2015,goodfellow2016}. 
	
	Given a deep neural network model that has an intermediate SSIM computational layer, the training procedure in a supervised learning setting can be formulated as follows:
	\begin{align}
	y_{i} &= f(x_{i},W ), ~~~ i \in \{1,...,N\},  \\
	l(f;x_{i},t_{i},W) &= l(f(x_{i},W),t_{i}),  ~~~ l \in \{1,...,N\},	 \\
	L &= \frac{1}{N} \sum_{i=1}^{N} l(f;x_{i},W ,t_{i}) + \sum_{j}{W_{j}^{2}} ,~~~ j \in \{1,...,W\},\\     
	\intertext{the weights are iteratively updated in the opposite direction of the gradient of the loss function as:}\\ 
	W_{k} &:= W_{k} - \eta \frac{\partial L}{\partial W_{k}}, ~~~ k \in \{1,...,K_{iterations}\},\\
	\intertext{for the SSIMLayer parameters, this update rule is changed to:}
	W_{k} &:= W_{k} - \eta \frac{\partial L}{\partial \text{SSIM}} \frac{\partial \text{SSIM}}{\partial y_{k}}, ~~~ y_k \subseteq W_{k} ,\\
	\intertext{the derivative expression $\frac{\partial L}{\partial \text{SSIM}}$ depends on the used global loss formula, for $\frac{\partial \text{SSIM}}{\partial y_{k}}$ of the local SSIM:}\\
	\text{SSIM}(x,y_{k}) &= \frac{(2\mu_{x}\mu_{y_{k}} + C_{1})(2\sigma_{xy_{k}} + C_{2})}{(\mu_{x}^{2} + \mu_{y_{k}}^{2} + C_{1})(\sigma_{x}^{2} + \sigma_{y_{k}}^{2} + C2)},\\
	\intertext{defining the following variables to simplify the expression:}
	\begin{split}
	A_{1} &= 2\mu_{x}\mu_{y_{k}} + C_{1},~~~ A_{2} = 2\sigma_{xy_{k}} + C_{2} \\
	B_{1} &= \mu_{x}^{2} + \mu_{y_{k}}^{2} + C_{1},~~~ B_{2} = \sigma_{x}^{2} + \sigma_{y_{k}}^{2} + C2
	\end{split},\\
	\intertext{then, the gradient of the SSIM is computed as in~\cite{wang2008maximum}: }\\
	\frac {\partial \text{SSIM}}{\partial{y_{k}}} &= \frac{2\bigg[[A_{1}B_{1}(B_{2}x-A_{2}y_{k})]+[B_{1}B_{2}(A_{2}-A{1})\mu_{x} \ast 1^{N_{p}}]+ [A_{1}A_{2}(B_{1}-B_{2})\mu_{y_{k}} \ast 1^{N_{p}}]\bigg]}{N_{p}B_{1}^{2}B_{2}^{2}},
	\end{align}
	\noindent where $W$ is the trainable weights, N is the number of training samples, $l(f;x_{i},t_{i},W)$ is the loss associated with a single training example, $L$ is the global loss over a single mini-batch, $\sum_{j}{W_{j}^{2}}$ is a regularisation term, $\eta$ is the learning rate, $x$ is the input, $y_{k} \subseteq W$ is the SSIM filter weights, $k$ is the update iteration index, $C_{1}, C_{2}$ are constants to ensure numerical stability and $1^{N
		_{p}} = [1, 1, 1, ... p]$ is a ones column vector of size $N_{p}$ pixels of a local patch.

	\section{Experimental Results}
	
	We evaluate the performance of the proposed SSIMLayer on the popular CIFAR-10 image classification dataset. CIFAR-10 consists of 50k training images and 10k testing images uniformly distributed among 10 classes. In the presented experiments, models are trained on the training set and evaluated on the test set. The evaluation protocol is that the SSIMLayer is injected into a deep learning architecture and compared to a convolutional one injected the same way in a plain convolutional architecture.
	
	\subsection{Training Details}
	\label{sec:ssim_train_details}
	Input images of dimensionality $32\times32\times3$ are randomly flipped on the horizontal axis for data augmentation. The stochastic gradient descent (SGD) algorithm with mini-batch size of 32 has been used for optimisation. The learning rate is fixed at 0.01 and the models are trained for up to 500 epochs.  A weight decay of 0.0001 and a momentum of 0.9 have been used. 
	\smallbreak
	Weight initialisation is of crucial importance to the convergence of deep neural networks. Several weight initialisation methods have been investigated. In order to converge, the SSIM weights are initialised from a standard normal distribution with zero mean and unit variance.

	\begin{figure}
		\centering
		\includegraphics[width=0.8\textwidth]{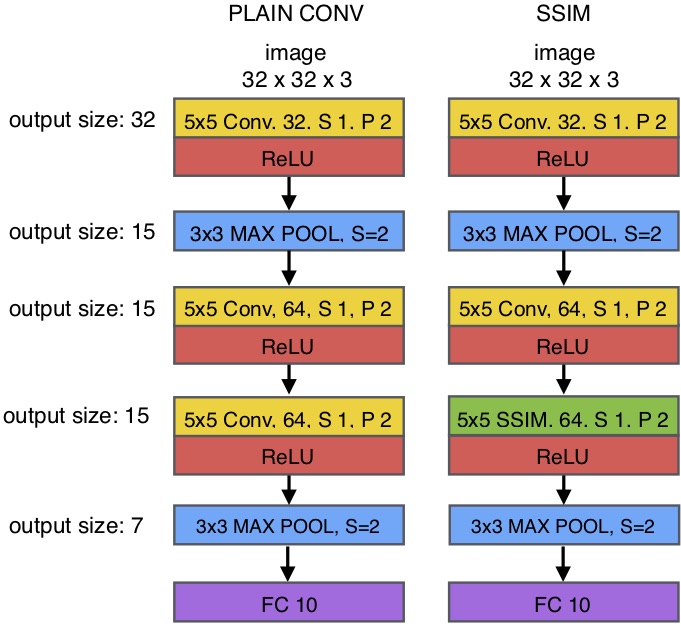}
		\caption[Network architectures used in evaluating the performance of the proposed SSIMLayer on CIFAR-10 dataset]{Network architectures used in evaluating the performance of the proposed SSIMLayer on CIFAR-10 dataset.  For a convolutional or SSIMLayer: (e.g, $5\times5~\text{Covn}, 32, \text{S}~1, \text{P}~2$) indicates a layer with 32 sliding window filters of size $5\times5$, stride $S = 1$ and padding $P=2$. This layer produces 32 feature activation maps that are passed through a nonlinear transformation using the rectified linear unit (ReLU).}
		\label{fig:conv_ssim_arch}
	\end{figure}
	
	\begin{figure}
		\centering
		\includegraphics[width=0.9\linewidth]{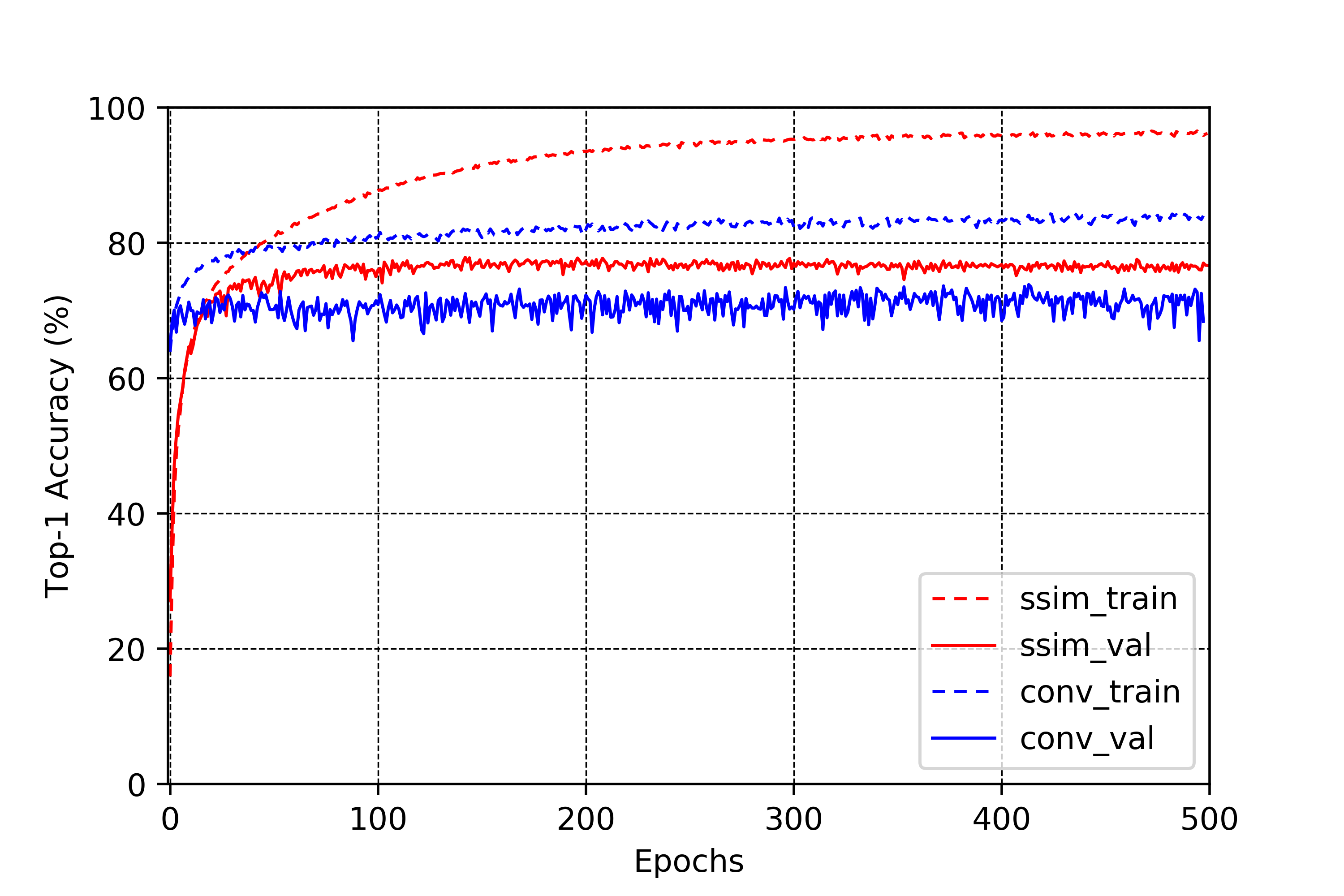}
		\caption[Training the SSIMLayer on CIFAR-10 dataset]{Evaluating the performance of the proposed SSIMLayer on the training and validation splits of CIFAR-10 dataset. The SSIM has been incorporated as an intermediate computational layer in a CNN architecture and compared with a similar plain convolutional architecture. Dashed curves represent training errors and solid lines denote validation errors.}
		\label{fig:cifar10_curves}
	\end{figure}

	\subsection{SSIM for Image Classification}
	This experiment demonstrates the effect of incorporating the SSIM as an intermediate computational layer in a deep learning architecture for CIFAR-10 image classification. The SSIMLayer operates as a high level feature extractor on top of the convolutional layers and before the output layer. This setting allows the SSIMLayer to build more meaningful structures from high level activations. Two architectures are compared: the first uses the SSIMLayer and the second is a plain convolutional architecture, as shown in Fig.~\ref{fig:conv_ssim_arch}. This is a controlled experiment where all training settings are fixed for both networks. 
	\smallbreak
	Figure~\ref{fig:cifar10_curves} shows convergence curves for both networks. As shown, the architecture that contains the SSIMLayer outperforms the plain convolutional network on the training and validation splits. It also demonstrates more confident behaviour on the validation split and higher capacity to accommodate the training data distribution. As detailed in Table\ref{table:ssim_cifar10}, the model containing the SSIMLayer achieves validation accuracy of 78.0 \% compared to 73.8 \% for the fully convolutional model. Further, the higher accuracy on the training set demonstrates high capacity to accommodate structures in the training set via modeling the human visual system capabilities in learning to extract high level structures in input feature maps.
	\smallbreak 
	
\begin{figure}[!h]
		\centering
		\begin{subfigure}[b]{0.450\textwidth}
			\centering
			\includegraphics[width=\textwidth]{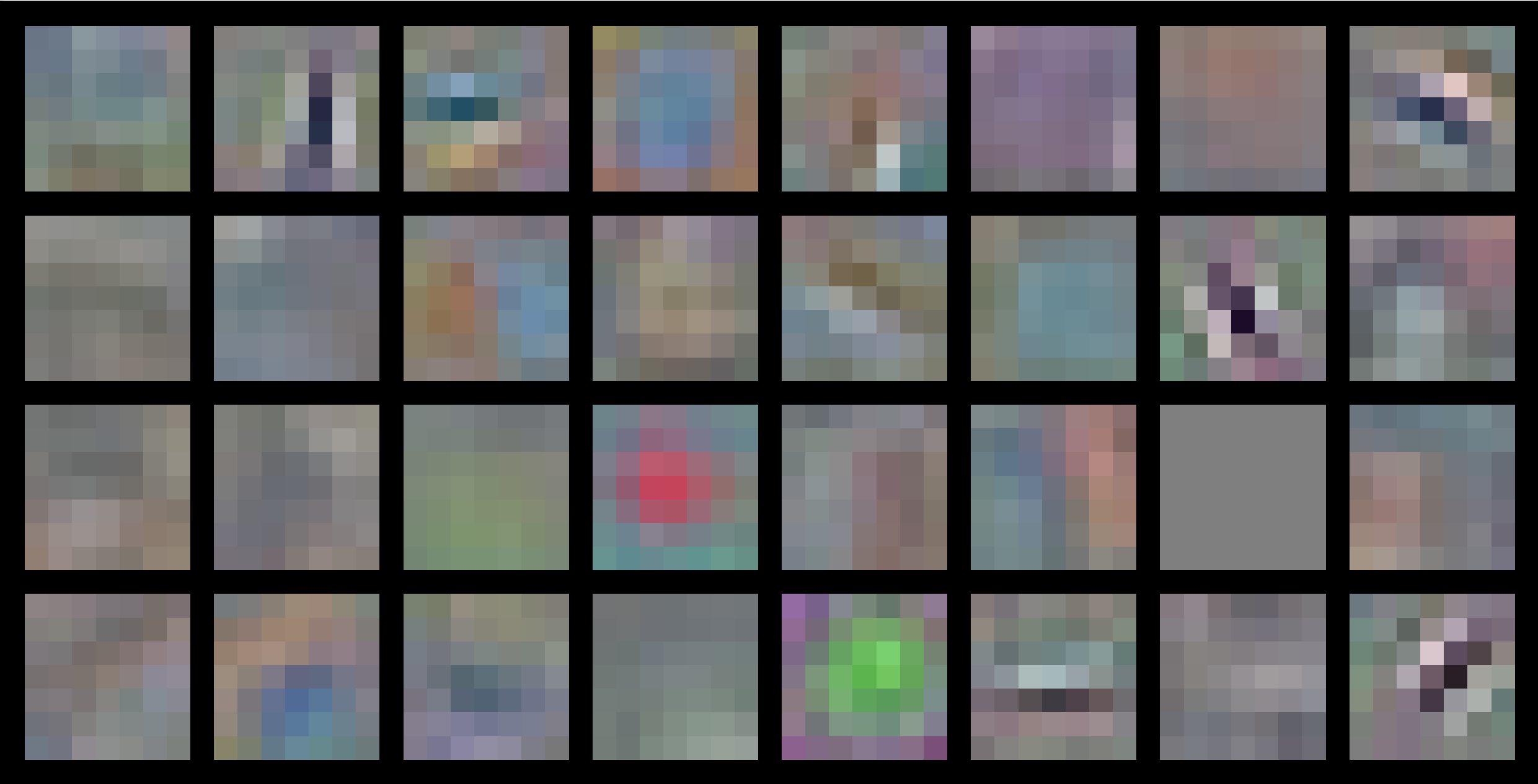}
			\caption{7x7 CONV Kernels}
			\label{fig:conv_filters}
		\end{subfigure}
		\hfill
		\begin{subfigure}[b]{0.45\textwidth}
			\centering
			\includegraphics[width=\textwidth]{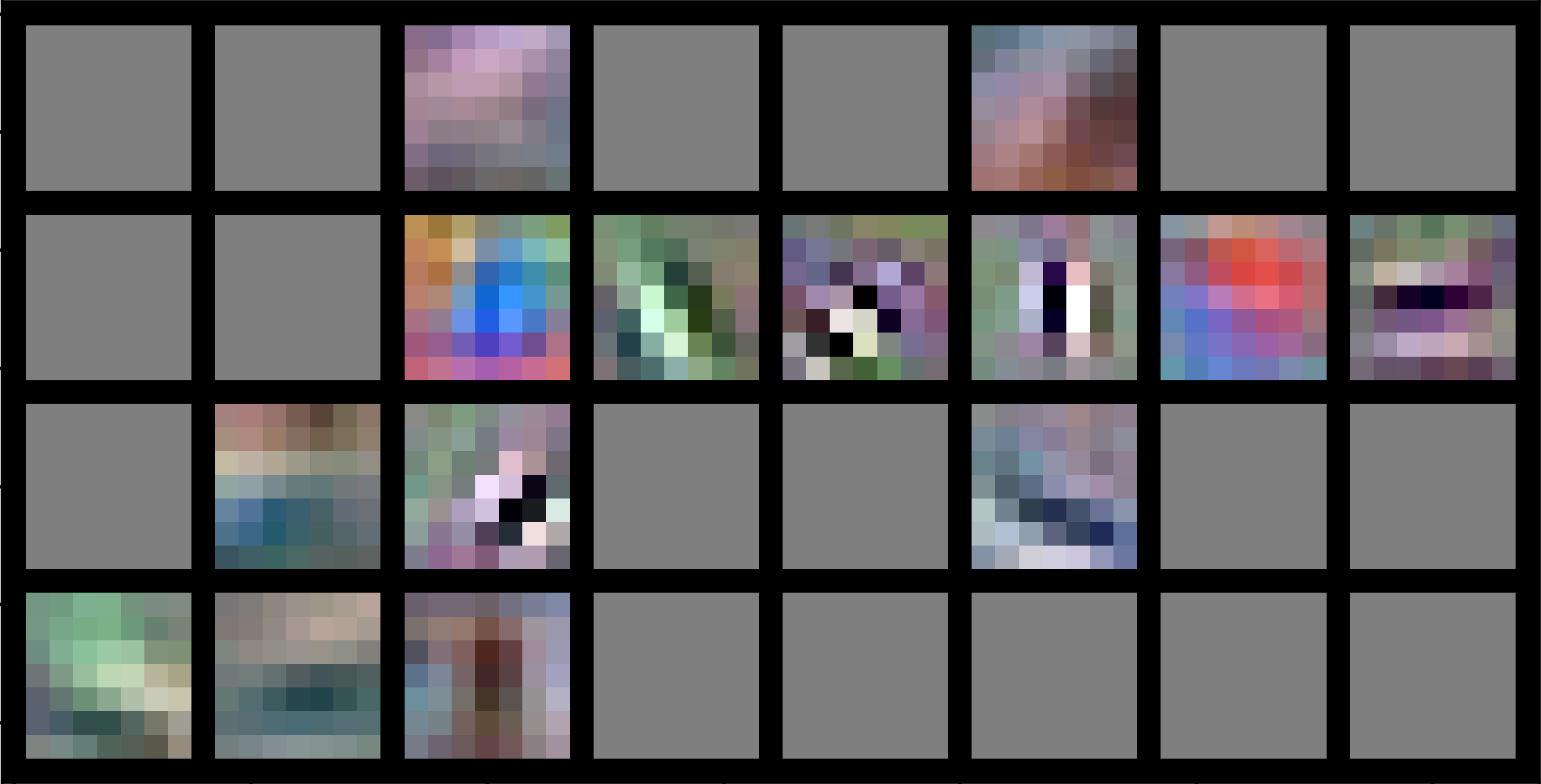}
			\caption{7x7 SSIM Kernels}
			\label{fig:ssim_filters}
		\end{subfigure}
		\caption{Visualisation of trained convolution and SSIMLayer kernels. The SSIM kernels represent the most common structures in the training dataset. As noted, a considerable number of SSIM filters have very small weights, $\text{norm} < 1$, this would suggest that a filter pruning stage is worth investigation.
		}
		\label{fig:filters}
	\end{figure}

	To visualise the trained low level filters, we trained two shallow models with the following architectures: \{7x7 SSIM - ReLU - MaxPOOL - 5x5 CONV - ReLU - MaxPOOL - FC\} and \{7x7 CONV - MaxPOOL - 5x5 CONV - ReLU - MaxPOOL - FC\} on CIFAR-10 dataset. 
	These models reported validation accuracy of 77.26 \% and 70.8 \%, respectively.
	Figure~\ref{fig:filters} depicts the low level trained convolution and SSIMLayer kernels. The SSIM kernels converged to the most common structures in the training dataset. However, as shown, a considerable number of SSIM filters have very small weights, $\text{norm} < 1$, this would suggest that a filter pruning stage is worth investigation.
	
	\begin{table}[t]
		\centering
		\captionsetup{justification=centering,labelsep=newline}
		\caption{Evaluating the performance of the proposed SSIMLayer on CIFAR-10 dataset}
		\begin{threeparttable}
			\setlength{\tabcolsep}{25pt}
			\begin{tabular}{l c c}
				\toprule
				\bf Model &  Training accuracy (\%) &  Validation accuracy (\%)  \\
				\midrule
				Plain Conv	&		84.9		 &	73.8		 \\
				Proposed SSIM			&		\textbf{96.5}		&	\textbf{78.0}	     \\
				\bottomrule    
			\end{tabular}
			\begin{tablenotes}
				\item
				\small
				The proposed SSIMLayer provides higher learning capacity than the traditional convolutional layer via modeling the human visual system capabilities in learning to extract high level structures in input feature maps. 
			\end{tablenotes}
		\end{threeparttable}		
		\label{table:ssim_cifar10}
	\end{table}	
	
	\begin{figure}
		\centering
		\includegraphics[width=0.8\textwidth]{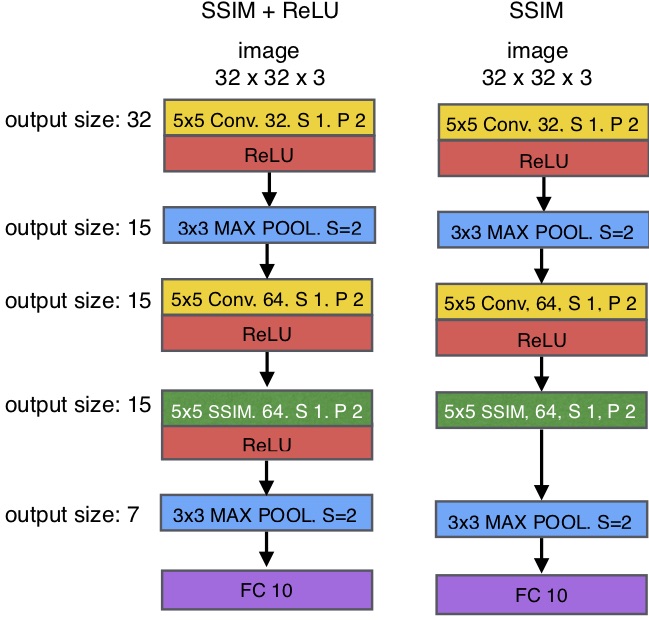}
		\caption{Network architectures used in evaluating the nonlinearity of the proposed SSIMLayer on CIFAR-10 dataset.}
		\label{fig:ssim_nonlinear_arch}
	\end{figure}

	\subsection{SSIM Nonlinearity}
	
\begin{figure}
		\centering
		\includegraphics[width=0.85\linewidth]{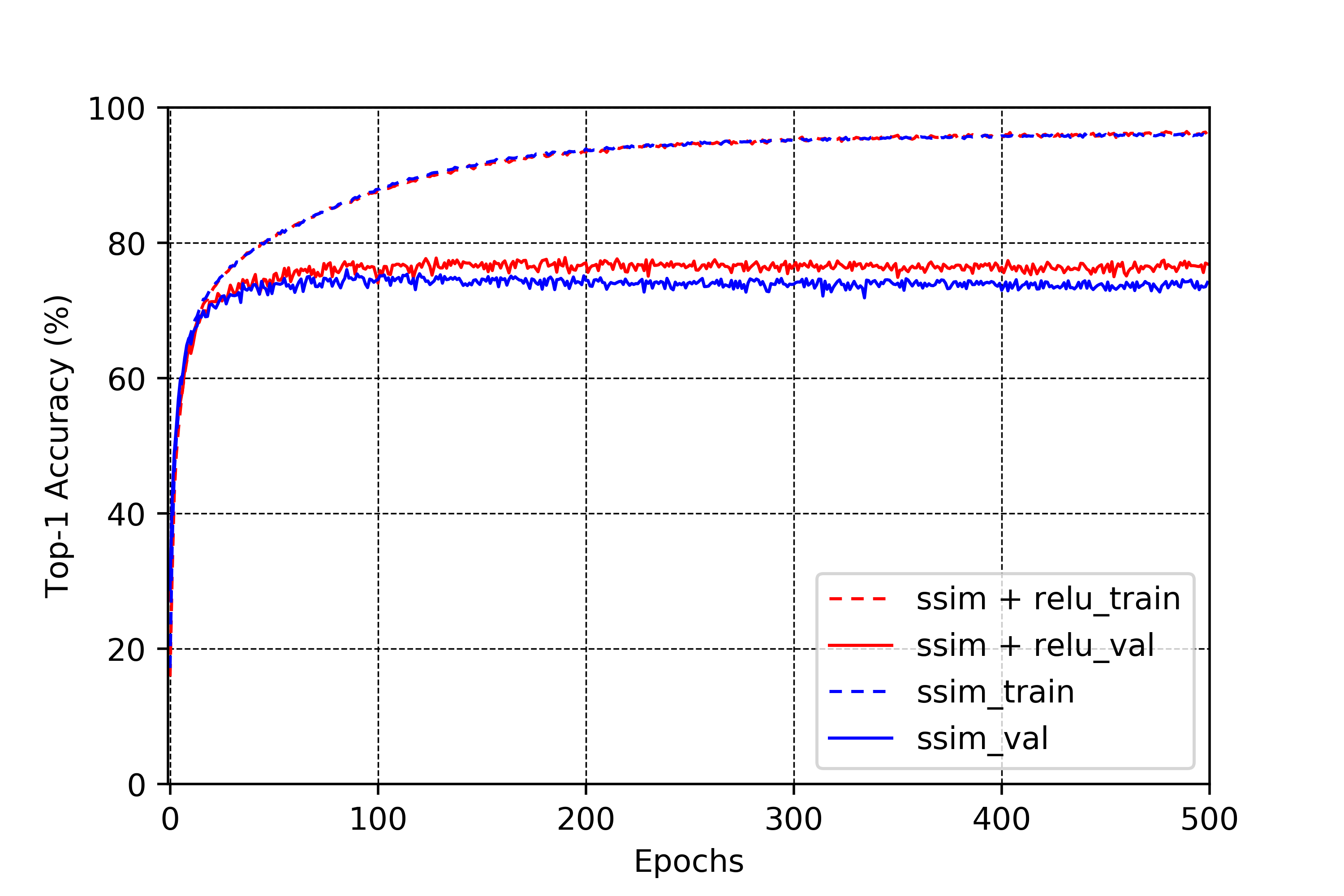}
		\caption[Evaluating the SSIM nonlinearity]{Evaluating the nonlinearity of the proposed SSIMLayer on the training and validation splits of CIFAR-10 dataset. Dashed curves represent training errors and solid lines denote validation errors. As shown, the effect of adding ReLU nonlinear transformation to the SSIMLayer is not significant.	
		}
		\label{fig:ssim_nonlinearity}
	\end{figure}

	The SSIM is a nonlinear operator that evaluates the degree of similarity between two input patches. In this case, a patch from the input feature maps and the sliding window filter. It produces a score that ranges from -1 to 1. The nonlinear nature of the SSIMLayer alleviates the need for a subsequent nonlinear activation function. This hypothesis has been tested via training the models shown in Fig.~\ref{fig:ssim_nonlinear_arch} on CIFAR-10 dataset. The main difference between both models is the existence of the ReLU nonlinear transfer function. The training convergence curves shown in Fig.~\ref{fig:ssim_nonlinearity} demonstrate that the effect of the added ReLU nonlinear transformation is not significant on training and validation sets.  Without nonlinear activation function, the SSIMLayer achieved maximum validation accuracy of 76.04 \% compared to 77.82 \% with ReLU added. Hence, the effect of adding ReLU nonlinear transformation to the SSIMLayer is not significant and can be compromised to reduce model complexity.

	\begin{figure}[h]
		\centering
		\includegraphics[width=0.85\linewidth]{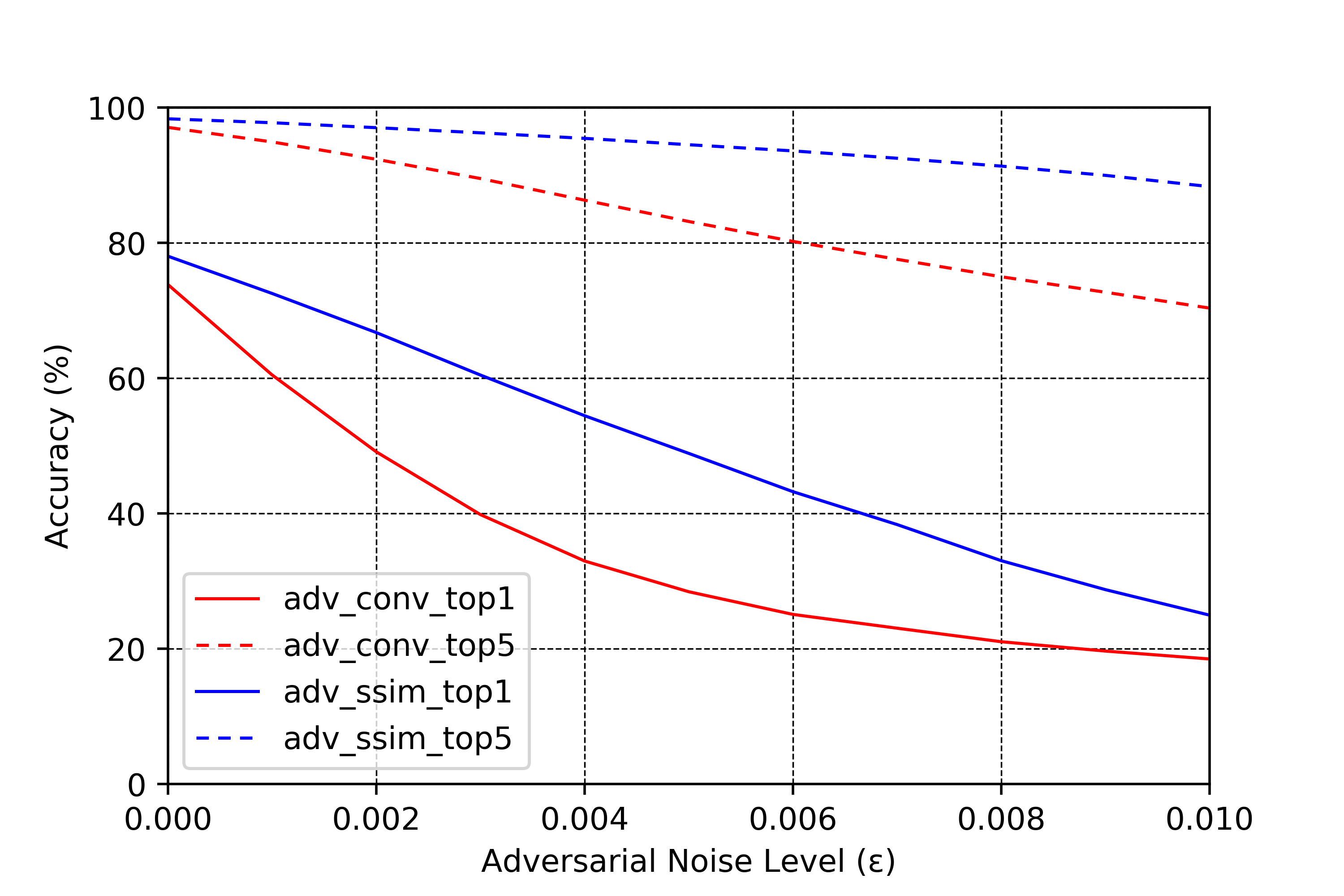}
		\caption[Comparing the effect of adversarial attacks on the proposed SSIMLayer.]{Comparing the effect of adding imperceptible perturbations to input images on the performance of a model with the SSIMLayer and a model with plain convolutional layers. 
		}
		\label{fig:adversarialssimconvcifar-10}
	\end{figure}
	
	\subsection{Robustness to Adversarial Examples}
	\label{sec:adv_attacks}
	
	The adversarial examples are defined as inputs that are perturbed using imperceptible noise~\cite{goodfellow2014explaining}. This noise causes machine learning models, not only the deep neural networks, to mis-classify these examples with high confidence. As it is the dominant for visual domain applications, the focus of this Chapter is to ensure more robustness for the deep ConvNet model via incorporating human perception capabilities to analyse structural information in the input. Goodfellow et al.~\cite{goodfellow2014explaining} have interpreted the adversarial attacks as a result of the linearity of deep neural network models in the high dimensional space. Since the proposed SSIMLayer is inherently nonlinear, it can hypothetically mitigate the severity of these attacks. To investigate the robustness of the proposed layer, the performance of the two models depicted in Fig.~\ref{fig:conv_ssim_arch} and trained on CIFAR-10 dataset is evaluated against adversarial examples generated using the fast gradient sign method (FGSM)~\cite{goodfellow2014explaining}. The FGSM algorithm generates adversarial examples by adding a fraction of the signed gradients of the cost function with respect to the input as follows. Given an input image $x$, the perturbed version is generated using the formula: 
	
	\begin{equation}
	adv_{x} = x + \epsilon \times sign(\nabla_{x}J(\theta, x, y))
	\end{equation}
	
	where $\epsilon$ is a fraction that controls the strength of added noise, $\theta$ is model parameters, $x$ is the input to the model, $y$ is the target label and $J(.)$ is the cost function used in training the ConvNet models. The $sign(.)$ function is defined as follows:
	
	\begin{equation}
	f(x)=
	\begin{cases}
	-1, & \text{if $\nabla_{x} < 0$}.\\
	0, & \text{if $\nabla_{x} = 0$}.\\
	1, & \text{if $\nabla_{x} > 0$}.		
	\end{cases}
	\end{equation}
	
	Figure~\ref{fig:adversarialssimconvcifar-10} compares the robustness of the proposed SSIMlayer with the traditional convolutional layer against adversarial attacks. Both models are initialised and trained with the same settings detailed in Section~\ref{sec:ssim_train_details}. As shown, the model that contains the SSIMLayer is more robust than the plain convolutional model on the training and validation splits of CIFAR-10 dataset at different adversarial noise levels $\epsilon$. For instance, at $\epsilon=0.007$ which represents the magnitude of the smallest bit of an 8 bit image~\cite{goodfellow2014explaining}, the SSIM model achieves a TOP-1 accuracy of 38.35\% and a TOP-5 accuracy of 92.5\% compared to 23.03\% and 77.56\% for the plain convolutional model on CIFAR-10 validation set. TOP-K represents the likelihood that the correct prediction is in the top K predictions made by the model. These results demonstrate the nonlinear nature of the SSIMLayer and its ability to extract structural information more independently of the effect of added noise. It is worth mentioning that, with $\epsilon>0.01$ on CIFAR-10 dataset, the images became unrecognisable and the misclassification rate of both models increased, especially, the model containing the SSIMLayer.

	\subsubsection{Higher Resolution CIFAR-10}
	\label{sec:adv_cifar_alike}
\begin{figure}[h]
		\centering
		\includegraphics[width=0.85\linewidth]{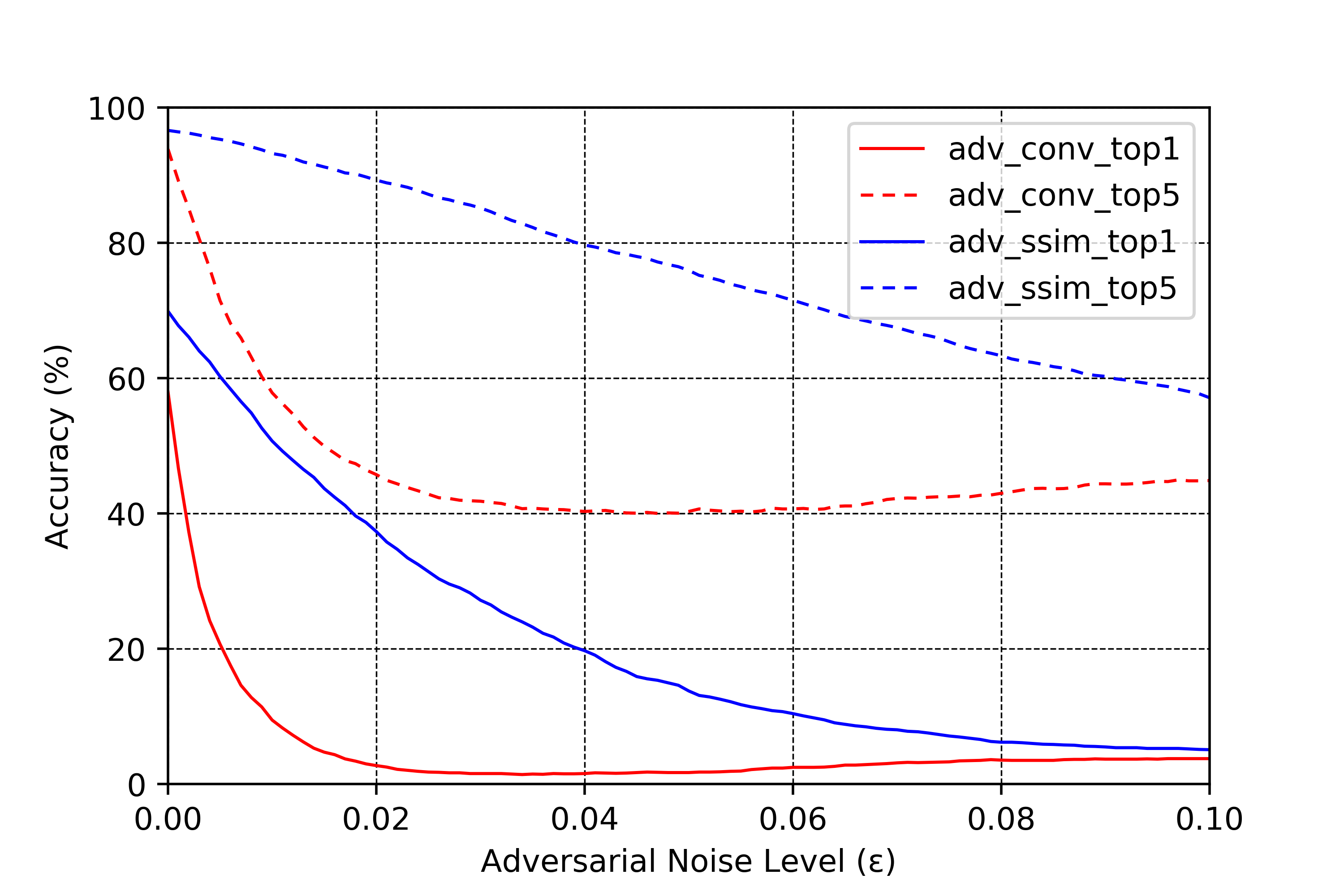}
		\caption[Comparing the effect of adversarial attacks on the proposed SSIMLayer.]{The SSIMLayer provides more robustness than the convolutional layer to adversarial attacks generated using the FGSM algorithm~\cite{goodfellow2014explaining} on ImageNet images of CIFAR-10 classes.
			
		}
		\label{fig:adversarialssimconvcifar-alike}
	\end{figure}

	In this section, we evaluate the performance of the SSIMLayer on ImageNet~\cite{imagenet2015} images of CIFAR-10 categories. We extracted a dataset of 15K images distributed uniformly among 10 classes, and split into 12K training and 3K validation splits. The models depicted in Fig.~\ref{fig:conv_ssim_arch} have been trained on this dataset. The images have been randomly resized and cropped to match CIFAR-10 dimensionality, normalised and randomly flipped on the horizontal axis for data augmentation. Table~\ref{table:ssim_cifar_alike} details training and validation accuracies of both models. The proposed SSIMLayer outperformed the traditional convolutional layer and demonstrated capabilities to build shallower models that can effectively learn from high quality and aggressively resized images. The SSIMLayer has further allowed the trained model to demonstrate robustness to more aggressive noise levels, as shown in Fig.~\ref{fig:adversarialssimconvcifar-alike}.  
	
	\begin{table}[t]
		\centering
		\captionsetup{justification=centering,labelsep=newline}
		\caption{Evaluating the performance of the proposed SSIMLayer on high resolution ImageNet images of CIFAR-10 classes}
		\begin{threeparttable}
			\setlength{\tabcolsep}{25pt}
			\begin{tabular}{l c c}
				\toprule
				\bf Model &  Training accuracy (\%) &  Validation accuracy (\%)  \\
				\midrule
				Plain Conv	&		57.51		 &	60.40		 \\
				Proposed SSIM			&		\textbf{73.35}		&	\textbf{70.0}	     \\
				\bottomrule    
			\end{tabular}
			\begin{tablenotes}
				\item
				\small
				The proposed SSIMLayer outperforms the traditional convolutional layer and demonstrates capabilities to build shallower models that can effectively learn from high quality and aggressively resized images. 
			\end{tablenotes}
		\end{threeparttable}		
		\label{table:ssim_cifar_alike}
	\end{table}

	\section{Conclusion}
	
	This paper proposed a new perceptually inspired computational layer (SSIMLayer) to the deep learning community. The proposed layer integrates the characteristics of the human visual system into deep convolutional neural networks. It allows the learning model to perceive structural information in input images. The proposed layer has a considerably higher learning capacity than the traditional convolutional layer allowing for building shallower and more efficient learning models. Further, it is inherently nonlinear and hence, it does not require subsequent nonlinear transformations. Experimental results demonstrate better convergence on CIFAR-10 dataset than the plain convolutional network. Also, the nonlinear nature of the SSIM operator allowed the deep model to be more robust to adversarial attacks.
	
	\section*{Acknowledgements}
	This research was fully supported by the Institute for Intelligent Systems Research and Innovation (IISRI) at Deakin University, Australia. 
	
	\bibliographystyle{IEEEtran}
	\bibliography{references}
	
\end{document}